\documentclass{article}

\usepackage{microtype}
\usepackage{graphicx}
\usepackage{subfigure}
\usepackage{booktabs} 

\usepackage{hyperref}

\usepackage[accepted]{icml2021}

\usepackage{amssymb,amsmath,amsthm,dsfont,bm,mathtools,amsfonts}
\usepackage{enumitem}
\setlist{topsep=10pt, partopsep=0pt, leftmargin=13pt} 
\usepackage{xcolor,color,graphicx,epsfig}

\newcommand{\loss}{{L}}

\newcommand{\smplst}{{\mathcal{S}}}

\newcommand{\reg}{{R}}
\newcommand{\dist}{{p}}

\newcommand{\dec}{{\Delta}}

\newcommand{\decapx}{{\tilde{\dec}}}

\newcommand{\rep}{{\phi}}

\newcommand{\layer}{{g}}
\newcommand{\ccp}{{\texttt{CCP}}}
\newcommand{\util}{{u}}

\newcommand{\temp}{{\tau}}
\newcommand{\mix}{{\gamma}}
\newcommand{\linvec}{{v}}
\newcommand{\tangent}{{T}}
\newcommand{\manifold}{{M}}
\newcommand{\pnlty}{{r}}
\newcommand{\cmix}[1]{{c_{\text{mix}}^{{#1}}}}
\newcommand{\clin}[1]{{c_{\text{lin}}^{{#1}}}}
\newcommand{\cquad}[1]{{c_{\text{quad}}^{{#1}}}}
\newcommand{\regutil}{{\reg_{\texttt{util}}}}
\newcommand{\regburden}{{\reg_{\texttt{burden}}}}
\newcommand{\regrecourse}{{\reg_{\texttt{recourse}}}}
\newcommand{\SERM}{{SERM}}

\newcommand{\spam}{{\texttt{spam}}}
\newcommand{\credit}{{\texttt{credit}}}
\newcommand{\finance}{{\texttt{financial distress}}}
\newcommand{\fraud}{{\texttt{fraud}}}

\newcommand{\red}[1]{{\leavevmode\color{red}{#1}}}
\newcommand{\blue}[1]{{\leavevmode\color{blue}{#1}}}
\newcommand{\green}[1]{{\leavevmode\color[RGB]{0,128,0}{#1}}}

\newcommand\todo[1]{\red{TODO: {#1}}}
\newcommand\tocite{\red{[CITE]}}

\newcommand\torefc[1]{\red{[REF: {#1}]}}
\newcommand\tocut[1]{\green{{#1}}}

\newcommand{\beq}{\begin{equation}}
\newcommand{\eeq}{\end{equation}}
\newcommand{\bal}{\begin{align}}
\newcommand{\eal}{\end{align}}
\newcommand\expect[2]{\mathbb{E}_{#1}{[ {#2} ]}}
\newcommand\prob[2]{\mathbb{P}_{#1}{\left[ {#2} \right]}}
\DeclareMathOperator*{\argmax}{argmax}

\DeclareMathOperator{\sign}{sign}

\newcommand{\1}[1]{\mathds{1}{\{{#1}\}}}

\newcommand{\naive}{{na\"{\i}ve}}

\newcommand{\X}{{\cal{X}}}

\newcommand{\x}{{\bm{x}}}
\newcommand{\Y}{{\cal{Y}}}

\newcommand{\Z}{{\cal{Z}}}
\newcommand{\C}{{\cal{C}}}

\newcommand{\R}{{\mathbb{R}}}

\newtheoremstyle{custom}
{4pt}
{4pt}
{\itshape}
{}
{\bfseries}
{:}
{.5em}
{}
\theoremstyle{custom}

\renewcommand{\tocut}[1]{{}}

\begin{document}
	
	\twocolumn[
	\icmltitle{Strategic Classification Made Practical}
	
	
	
	\icmlsetsymbol{equal}{*}
	
	\begin{icmlauthorlist}
		\icmlauthor{Sagi Levanon}{technion}
		\icmlauthor{Nir Rosenfeld}{technion}
	\end{icmlauthorlist}
	
	\icmlaffiliation{technion}{Department of Computer Science, Technion - Israel Institute of Technology}
	
	\icmlcorrespondingauthor{Nir Rosenfeld}{nirr@cs.technion.ac.il}
	
	\icmlkeywords{Machine Learning, ICML, strategic classification, differentiable optimizaiton}
	
	\vskip 0.3in
	]
	
	
	
	\printAffiliationsAndNotice{}  
	
	\begin{abstract}


Strategic classification regards the problem of learning in settings
where users can strategically modify their features to improve outcomes.
This setting applies broadly and has received much recent attention.
But despite its practical significance,
work in this space has so far been predominantly theoretical.
In this paper
we present a learning framework for strategic classification that is \emph{practical}.
Our approach directly minimizes the ``strategic'' empirical risk,
achieved by differentiating through the strategic response of users.
This provides flexibility that allows us to extend beyond the original problem formulation
and towards more realistic learning scenarios.
A series of experiments demonstrates the effectiveness of our approach
on various learning settings.

%
%
%
%

	\end{abstract}
	

\section{Introduction}


Across a multitude of domains---from loan approval to online dating to hiring and admissions---predictive machine learning models are becoming imperative for informing decisions that affect the lives of humans.
But when people benefit from certain predictive outcomes,
they are prone to act strategically to improve those outcomes.
This 
has raised awareness as to the idea that standard learning algorithms may not be robust to such behavior,
and there is a growing recognition as to the prevalence of this phenomena.
Given the breadth of domains in which strategic user behavior is likely,
practical tools for learning in strategic settings are of considerable importance.


In this paper we study practical aspects of learning in the setting of
\emph{strategic classification} \citep{BrucknerS11,hardt2016strategic}.
In this problem, users respond to a published classifier
by strategically modifying their features (at some cost) to improve their predicted outcomes.
As a concrete example, 
consider a bank offering loans.
The bank would like to approve a loan only if it is likely to be returned,
and to aid in decision-making, the bank trains a classifier to predict loan returns.
Applicants, however, would like their requests to be approved---regardless of their actual credibility.
To promote their interests, they can modify their applications (at some cost)
to best align with the bank's classification rule.
From the bank's perspective, such modifications can invalidate model predictions,
and the primary goal in strategic classification is to design learning algorithms
that are robust to this form of `gaming'.

Note that gaming is caused \emph{by the model itself},
indirectly through how it shapes user incentives, and to its detriment.
In this sense, strategic classification exemplifies how machine learning can be amenable to 
Goodhart's Law, a principle in policymaking 
stating that ``when a measure becomes a target, it ceases to be a good measure''. 
Strategic classification thus succinctly captures a natural form of tension
that arises between a learning-based system and its users.
There has been much recent work on this topic,
studying aspects such as generalization \citep{sundaram2020pac,zhang2021incentive},
equilibrium and dynamics \citep{PZM20,brown2020performative,izzo2021learn,miller2021outside},
online learning \citep{DongRSWW18,CLP19,ABBN19},
causality and decision outcomes \citep{KleinbergR19,lookahead,shavit2020causal,BLWZ20,miller2020strategic},
transparency \citep{darksc,BechavodPZWZ21},
and social perspectives \citep{HuIV19,MilliMDH19,chen2020strategic}.


But despite this flurry of recent work,
algorithms for strategic classification are not in widespread use.
The key reason for this is that work in this space has been predominantly theoretical.
This has several practical implications.
First,
for mathematical tractability, strong assumptions are made
(e.g., that the cost function is fixed and known).
But the robustness of these methods to violations of their assumptions is not well understood.
Second,
methods tend to be crafted for very particular learning settings,
and do not easily extend beyond the narrow context in which they are originally studied.\tocut{\footnote{\citet{hardt2016strategic}, for example,
focus on realizable binary classification under certain ``decomposable'' cost functions,
and their algorithm relies heavily on this assumption of decomposability and in a very explicit way
(the cost function is used to construct the output classifier).}}
Third,
currently available algorithms lag far behind recent advances in machine learning methodology;
they are prone to issues of scale, expressivity, and runtime,
and lack the flexibility and modularity that current approaches readily provide
(e.g., the ability to seamlessly ``add a layer'' or change the loss function).
Combined,
the above limitations indicate a clear need for an approach to learning
in strategic classification that is effective and practical.
Our work aims to take a first step towards addressing this need.

The core idea of our approach is to encode into the learning objective
a \emph{response mapping} that models how users respond to a given classification rule.
By anticipating how inputs will be strategically modified under a given model,
our approach optimizes directly for predictive performance under strategic behavior.
We refer to this as \emph{strategic empirical risk minimization}, or SERM. 
The challenge in optimizing the \SERM\ objective is  
that models of user response typically involve an argmax operator,
which can be non-differentiable and even discontinuous.
Our solution is to replace the argmax operator with a differentiable proxy,
and for this we draw on recent advances in differentiable optimization solvers
\citep{amos2017optnet,djolonga2017differentiable,agrawal2019differentiable,agrawal2019differentiating,berthet2020learning,tan2020learning,agrawal2020differentiating}
and adapt them to our purpose.
The resulting \emph{strategic response layer} provides the main building-block of our framework.


Using the flexibility our framework provides,
we propose and showcase multiple ways in which the framework
can extend beyond the original formulation of strategic classification
and towards more realistic learning scenarios.
We make use of the modular nature of our approach and the flexibility it provides
to explore various extensions aimed at addressing potential practical concerns.
These include:
supporting complex predictive models (e.g., recursive neural networks),
supporting structured cost functions (e.g., constraining movement to a manifold),
and
relaxing the assumption of a fixed and known cost function
(we handle adjustable and unknown costs).

%
%
%

The primary goal in strategic classification is to learn strategically-robust predictive models,
but several works have raised concern as to the adverse social outcomes this may entail
\citep{MilliMDH19,HuIV19,chen2020strategic}.
This may not be surprising given that learning focuses entirely on optimizing predictive accuracy.
To address these concerns,
here we argue for a broader perspective that considers the trade-off between
system and user interests.
Borrowing from welfare economics,
we take the perspective of a `social planner' tasked with balancing between these interests,
and show how our approach can extend to target any operating point along the pareto front.
To do this, we cast strategic classification as a problem of \emph{model selection},
and propose novel forms of regularization that promote favorable social outcomes
for various notions of `social good'.
Because strategic classification is not a zero-sum game,
the incentives of the system and of its users are not entirely antagonistic.
As we show, this permits much social benefit to be gained at only a small cost in accuracy.

In summary, our paper makes the following contributions:
\begin{itemize}
\item 
\textbf{Practical framework.} We propose a novel learning framework for strategic classification
that is practical, effective, and flexible.
Our approach allows to differentiate through strategic user responses, thus permitting end-to-end training.

\item 
\textbf{Flexible modeling.}
We show how the flexibility of our approach allows for 
learning in diverse strategic settings.
We effectively apply our approach to multiple such settings of practical interest.

\item 
\textbf{Socially-aware learning.}
We propose several forms of regularization that encourage learned models to promote
favorable social outcomes.
By capitalizing on certain structural aspects of the problem,
our regularization effectively balances between system and user interests.
\end{itemize}	

We conduct a series of experiments demonstrating the effectiveness of our approach.
With respect to the above points,
each of our experiments is designed to study a different practical aspect of learning.
The experiments cover a range of learning environments using real and synthetic data.
Our results show that learning in strategic classification can be practical,
effective, and socially responsible.

One of our main goals in this paper is to motivate and support future
empirical research on strategic classification,
Towards this end, we make publicly available a code repository with a
flexible implementation of our approach,
designed to support a wide range of strategic learning settings.
Code can be found at
\url{https://github.com/SagiLevanon1/scmp}.

\section{Related Work}


The literature on strategic classification is growing at a rapid pace.
Various formulations of the problem were studied in earlier works
\citep{bruckner2009nash,bruckner2012static,grosshans2013bayesian},
but most recent works adopt the core setup of \citet{hardt2016strategic}, as we do here.
Research in this space has mostly been oriented towards theory,
with recent work introducing notions similar to SERM and extending PAC theory to this setting
\citep{sundaram2020pac,zhang2021incentive}.
We complement these by placing emphasis on practical aspects of learning.

Several papers consider the social impacts of strategic classification.
\citet{MilliMDH19} study the social burden imposed by optimizing for accuracy. 
\citet{chen2020strategic} study the connection between strategically-aware learning and recourse.
\citet{HuIV19} focus on fairness and show how classifiers can induce inequitable modification costs
that affect utility.
Our work ties these together, providing means to \emph{control} the tradeoff between
classifier's accuracy and the social outcomes it induces through regularization (see Sec. \ref{sec:reg}).

A parallel line of work studies how learned models affect actual (rather than predictive) outcomes.
Some works analyze how models should promote users to invest effort effectively \citep{KleinbergR19,AlonDPTT20},
while others tie learning to the underlying casual mechanisms of the environment
\citep{PZM20,BLWZ20,shavit2020causal,miller2020strategic}. 
We remain within the original, purely predictive problem formulation,
but view the extension of our approach to such settings to be intriguing as future work.

\section{Method} \label{sec:method}

\paragraph{Learning setup.} \label{sec:setup}
Denote by $x \in \X \subseteq \R^d$ features representing user attributes
(e.g., a loan application profile),
and by $y \in \Y = \{-1,1\}$ their corresponding labels (e.g., loan returned or not).
Let $\dist(x,y)$ be a joint distribution over non-strategic features and labels.
The primary goal in learning is to find a classifier
$h : \X \rightarrow \Y$ from a class $H$
that achieves high expected accuracy.
For this, we assume access to a sample set $\smplst=\{(x_i,y_i)\}_{i=1}^m$
sampled i.i.d. from $\dist(x,y)$, on which we train.
At test time, however,
$h$ is evaluated on data 
that is prone to modification by users. 
In strategic classification,
users modify their features using the \emph{response mapping}:
\begin{equation}
\label{eq:best_response}
\dec_h(x) \triangleq \argmax_{x' \in \X} h(x') - c(x,x')
\end{equation}
where $c$ is a known cost function.
Test data includes pairs $(\dec_h(x),y)$ where $(x,y)\sim \dist$,
and the goal in learning is to optimize predictive accuracy under this induced distribution.


As is common, the classifiers we consider will be based on score functions $f:\X\rightarrow\R$
via the decision rule $h_f(x)=\sign(f(x))$,
and learning will be concerned with optimizing
over a class of parametrized score functions $F$.
We write $\dec_f$ to mean $\dec_{h_f}$,
and for clarity, omit the notational dependence of the classifier
$h_f$ on $f$ when clear from context.

\paragraph{Game-theoretic formulation.} \label{sec:game}
Strategic classification can be formulated as a Stackelberg game
between two players---the \emph{system} and a population of \emph{users}.
First, the system learns from $\smplst$ a classifier $h$. 
Then, given $h$, users respond via $\dec_h$.
The payoffs are:
\begin{align}
\text{{System}:} & \quad
\prob{}{y = h(\dec_h(x))}, \label{eq:system_payoff} \\
\text{{Users}:} & \quad
\expect{}{h(\dec_h(x)) - c(x,\dec_h(x))} \label{eq:users_payoff}
\end{align}
Payoff to the system (Eq. \eqref{eq:system_payoff}) 
is the probability of classifying manipulated points correctly.
Payoff to the users (as a collective) is their expected utility (Eq. \eqref{eq:users_payoff}),
for which $\dec$ as defined in Eq. \eqref{eq:best_response} is a best-response. 



\subsection{Strategic Empirical Risk Minimization} \label{sec:serm}


A \naive\ approach would be to train a classifier to predict well on the (non-manipulated) input data,
and use it at test time on manipulated data.
The caveat in this approach is that strategic behavior causes
a discrepancy between the (marginal) input distributions at train and test time.
Strategic classification therefore introduces a form of
\emph{distribution shift} \citep{quionero2009dataset},
but with the unique property that shift is 
determined by the predictive model itself,
albeit indirectly through its effect on user responses. 

Our approach will be to account for this shift by directly optimizing
for the induced distribution.
Noting that the system's payoff in Eq. \eqref{eq:system_payoff} can be rewritten as
$1-\expect{}{\1{y \neq h(\dec_h(x))}}$,
we set our objective to 
the empirical loss over the \emph{strategically-modified} training set:
\begin{equation}
\label{eq:ERM}
\min_{f \in F} \sum_{i=1}^m \loss(\dec_f(x_i), y_i, f) + \lambda \reg(f) 
\end{equation}
where $\loss(z, y, f) = \1{y \neq h_f(z)}$
and $\reg$ is an optional regularizer.
In practice we replace $\loss$ with a tractable surrogate (e.g., binary cross-entropy),
and we will return to the role of regularization in Sec. \ref{sec:reg}.
We refer to optimizing Eq. \eqref{eq:ERM} as
\emph{strategic empirical risk minimization} (\SERM).
Note that $f$ plays a dual role in the objective:
it determines how inputs are modified (via $\dec_f$)
and how predictions are made on those modified inputs (via $h_f$).

\subsection{Differentiating through strategic responses} \label{sec:diff_through_response}

A natural way to approach the optimization of Eq. \eqref{eq:ERM} is using gradient methods.
The challenge in this approach is that $\dec$ is an argmax operator and can therefore
be non-differentiable. 
Our solution to this will be to use a differentiable proxy for $\dec$,
drawing inspiration from recent advances in \emph{differentibale optimization solvers}.
Such solvers map parametrized optimization problems to their (approximately) optimal solutions
in a manner that is amenable to differentiation,
and so can be used as optimization ``layers'' integrated into neural architectures.

Our approach will be to
implement the response mapping $\dec$ as a differentiable optimization layer.
In particular, we make use of \emph{convex optimization layers} \citep{agrawal2019differentiable},
but since we seek to maximize, we will construct \emph{concave} layers.
A concave optimization layer $\layer(\theta)$ maps concave optimization problem instances
to their argmax:
the input to the layer, $\theta$, defines the ``parameters'' (and hence the instance)
of a template optimization problem,
and the output of the layer is the solution under this parameterization.\tocut{\footnote{For a (convex) example, if $\layer(\theta)$ encodes a shortest paths problem, then $\theta$ can determine edge weights,
	and $y=\layer(\theta)$ encodes a shortest path under these weights
	as determined by a linear programming solution.}}
In our model, $\layer$ will play the role of $\Delta$, and $\theta$ will include
features $x$ and the learnable parameters of $f$.

The response mapping $\dec$ as defined in Eq. \eqref{eq:best_response} is not concave,
and to apply concave optimization, we construct a concave proxy, denoted $\decapx$, as follows.
To begin, assume for simplicity that $h$ is linear, i.e., $h_w(x)=\sign f_w(x)$ with
$f_w(x)= w^\top x + b$\tocut{ (this assumption will be relaxed later)}.
%
Next, since $\sign$ is discontinuous, we replace it with a smooth sigmoid $\sigma^{*}$
of the following form:
\begin{equation*}
\sigma^{*}_{\tau}(z) = 
\frac{1}{2} \sqrt{\left(\temp^{-1} z+1\right)^{2}+1} - 
\frac{1}{2}\sqrt{\left(\temp^{-1} z-1\right)^{2}+1}
\end{equation*}
Here $\temp$ is a temperature parameter: as $\tau$ decreases, $\sigma^{*}$ approaches $\sign$.
Our particular choice of sigmoid follows from the fact that $\sigma^{*}$ can be written as a sum of convex
and concave functions.
This motivates our final step,
which is to apply the convex-concave procedure (CCP) \citep{yuille2003concave}.
CCP is an approach to solving convex-concave optimization problems by
iterating through a sequence of concave-relaxed problems,
a process that guarantees convergence to local maxima.
We focus on convex costs (e.g., linear or quadratic) so that
CCP can be applied to the entire response function.\footnote{Most of the literature considers convex costs
	(and linear classifiers). This includes the experimental setting of \citet{hardt2016strategic}.}
Using CCP, we obtain reliable concave proxies
of responses at each input.
This gives us our differentiable proxy of the response mapping,
which we refer to as a \emph{strategic response layer}, defined as:
\begin{equation}
\label{eq:decapx}
\decapx(x) = \argmax_{x' \in \X} \ccp(\sigma^{*}(w^\top x' + b)) - c(x,x')
\end{equation}
Here $\ccp$ denotes the concave proxy obtained at the last iteration of the procedure.


To compute the forward pass, we run the CCP procedure
using any standard off-shelf convex solver,
and compute the argmax w.r.t. the final proxy.\footnote{There are many such tools available, and for reasonably-sized inputs,
	the computation overhead is small.}
For the backward pass, we plug the final proxy into the differentiable
convex solver of \citet{agrawal2019differentiable}
to get gradients for $\decapx$ w.r.t. $w,b$.



\subsection{Extensions} \label{sec:extensions}
\subsubsection{Nonlinear classifiers} \label{sec:nonlinear}
The concave solver can handle only score functions $f$ that are linear in the optimization variables.
But $f$ need not be linear in the input features $x$;
rather, it can be linear in any high-dimensional representation of the inputs, $z=\rep(x)$, $z \in \Z$,
such as those obtained from the final hidden layers of neural networks.
This holds as long as both $f$ and $c$ are defined over this representation
(i.e., points move in representation-space).\footnote{Cost over representations are sensible,
	for example, when latent dimensions correspond to meaningful and manipulable real-world
	properties (e.g., \citet{infogan}).
}
The response mapping becomes: 
\begin{equation} \label{eq:br_rep}
\dec(x) = \argmax_{z' \in \Z} \sigma(w^\top z' + b) - c(\rep(x),z')
\end{equation}
More generally, $f$ must be linear in variables over which the cost function $c$ is defined.
For example, if $c$ applies only to a subset of the original features
(for example, if only some features are manipulable),
then $f$ must be linear those features, 
but can be non-linear in all other features.
Concretely, if $x=(x_{\mathtt{manip}},x_{\mathtt{non}})$ where $x_{\mathtt{manip}}\in \R^{d_1}$
and $x_{\mathtt{non}}\in \R^{d_2}$ are the manipulable and non-manipulable features,
respectively, 
then our framework supports the following non-linear representational structure:
\begin{equation*} \label{eq:rep_manip}
\dec(x) = \argmax_{x' \in \R^{d_1}} \sigma(w^\top x' + v^\top \phi(x_{\mathtt{non}}) + b) - c(x_{\mathtt{manip}},x')
\end{equation*}

\subsubsection{Flexible costs} \label{sec:flexibility}
The core setting of strategic classification assumes costs are fixed.
But in some settings, it is reasonable to assume that the system has some control over the cost function.\footnote{For example,
consider a bank offering loans, and assume one of the features is the number of credit cards a user has. If the bank itself issues credit cards, then by determining policies related to credit cards (e.g., eligibility criteria, commissions, rates), the bank achieves some control over costs.}
We model this as allowing the system to modify an initial cost function $c_0$
to some other cost $c \in \C$ from the class $\C$, with this incurring a penalty of $\pnlty(c_0,c)$.
The goal is now to jointly learn the classifier \emph{and} the modified cost.
Denoting by $\dec_f^c$ the response mapping for classifier $f$ and cost function $c$,
we can extend the learning objective as follows:
\begin{equation} \label{eq:flex_obj}
\min_{c \in \C} \min_{f \in F} \sum_{i=1}^m \loss(\dec_f^c(x_i), y_i, f) 
+ \pnlty(c_0,c)
\end{equation}
By utilizing its flexibility in choosing $c$,
in this setting 
the system has the capacity to obtain better predictive performance than when optimizing
the objective in Eq. \eqref{eq:ERM}.

\subsubsection{Unknown costs} \label{sec:robustness}
Strategic classification assumes that the cost function is known,
but this may not hold in practice.
Here we consider a setting in which the system does not know the true cost $c^*$,
but has a reasonable estimate $c_0$.
We model the system as believing that $c^*$ lies in some set $C \subseteq \C$ which includes $c_0$
as well as nearby points,
and which we view as a design parameter chosen by the learner.
To learn in this setting, we propose a worst-case approach in which the system aims to perform well
simultaneously on \emph{all} $c \in C$.
We propose the following minmax objective:
\begin{equation} \label{eq:rob_obj}
\min_{f \in F} \max_{c \in C} \sum_{i=1}^m \loss(\dec_f^c(x_i), y_i, f)
\end{equation}

\subsubsection{Moving on a manifold} \label{sec:manifold}
The Manifold Hypothesis is a convention stating that high-dimensional data
tend to lie on or near a low-dimensional manifold. 
Typically the manifold is unknown; but even if it is known, 
common cost functions tend to permit arbitrary movement and so cannot account for this structure.
Here we show how our approach can incorporate (approximate) manifold constraints into learning.
We begin by learning a \emph{manifold bundle} $(\manifold, \tangent)$
composed of a manifold model $\manifold$
and a tangent function $\tangent(x)$ that returns the subspace tangent to $\manifold$ at $x$.\footnote{Many tools exist for learning bundles; e.g., \citet{rifai2011manifold}.} 
Since tangents $T(x)$ are linear objects that provide a first-order approximation
to the manifold at $x$, our approach is to add them as linear constraints to the
optimization of the response mapping (i.e., the argmax in $\dec$ is taken over $x' \in T(x)$).
This ensure points move only on the tangent,
thus approximating movement on the manifold.

%



\subsection{Regularizing for social good} \label{sec:reg}

Our discussion of strategic classification thus far has been focused on a single objective:
predicting accurately in the face of strategic gaming.
This promotes a clear interest of the system, but
neglects to account for the effects of learning users.
Returning to our loans example, note that any predictor 
inevitably determines the degree of \emph{recourse},
defined as the ability of users that are denied a service (e.g., a loan)
to take reasonable action to reverse this decision \citep{ustun2019actionable,gupta2019equalizing,joshi2019towards,chen2020strategic,karimi2020algorithmic}.
Recourse is clearly beneficial to users,
but in many cases, its facilitation is also beneficial to the system (for discussion see
\citet{ustun2019actionable,venkatasubramanian2020philosophical,karimi2020survey}).

In this section we show how our framework can be used to train models
that are both accurate \emph{and} promote favorable social outcomes.
Relying on the observation that different models can induce very different social outcomes \citep{heidari2019long},
we cast learning as a problem of \emph{model selection},
where the selection criterion
reflects some notion of `social good' (e.g., recourse).
Model selection is implemented through regularization 
and below we present novel forms of data-dependent regularizers $\reg(f;\smplst)$ targeting
various notions of social good from the literature.

Intuitively, we expect regularization to be useful because
\emph{strategic classification is not a zero-sum game}.
In other words, predictive models that provide similar payoff to the system
may differ considerably in their payoff to users;
we argue that the system has the freedom, as well as the responsibility,
to carefully choose between these.
Viewing learning from the perspective of a `social planner'
interested in balancing between system and user interests,
our regularization approach provides the means to achieve good balance.
Varying the amount of regularization $\lambda$
provides solutions along the Pareto front,
with 
$\lambda=0$ corresponding to the strategic classification equilibrium in Eq. \eqref{eq:system_payoff}.

\paragraph{Expected utility.}
Since the payoff to users  in Eq. \eqref{eq:users_payoff} is given by their gained \emph{utility},
$\util_\dec(x) = h(\dec(x))-c(x,\dec(x))$,
a straightforward notion of social good is their \emph{expected utility}, $\expect{}{\util_\dec(x)}$.
Note that $\dec$ is a best-response but it is utility-optimal \emph{relative to $h$},
and it is easy to construct an example showing that under the accuracy-optimal predictor
utility can be arbitrarily low.\footnote{
Consider $d=1$, with points $(-\epsilon,-1)$
and $(\epsilon,1)$. Let $c(x,x')=|x-x'|$. The classifier $h(x)=\1{x \ge 2}$ causes all positive points (and only positive points) to move. The payoff to negative points is $-1$ and the payoff to positive points is $-1+\epsilon$.}
To encourage models that provide users with high utility, we set:
\begin{equation}
\label{eq:reg_exputil}
\regutil(f;\smplst) = - \sum_{i=1}^m \util_\dec(x_i)
\end{equation}
and plug into the learning objective in \eqref{eq:ERM},
where in practice we again replace $\dec$ with $\decapx$ and optimize as in
Sec. \ref{sec:diff_through_response}.


\paragraph{Social burden.}
In their paper on the social cost of strategic classification,
\citet{MilliMDH19} study \emph{social burden},
defined to be the minimum cost a positively-labeled user
must incur in order to be classified correctly.
Within our framework, we can regularize for social burden using:
\begin{equation}
\label{eq:reg_burden}
\regburden(f;\smplst) = \sum_{i : y_i=1} \min_{x' : f(x') \ge 0} c(x, x')
\end{equation}
which we can also implement as a convex optimization layer for linear $f$ and convex $c$
(the constraint is linear in $x,w$).

\paragraph{Recourse.}

Recourse refers to the capacity of a user who is denied a service
to restore approval through reasonable action (in our case, low-cost feature modification).
Since $\dec$ is a best-response, we say a user with $h(x)=-1$ is \emph{granted recourse} if $h(\dec_h(x))=1$.
The random variable negating this condition is $\1{h(x)=-1 \, \land \, h(\dec_h(x))=-1}$,
and we regularize using its smoothed approximation:
\begin{equation*}
\label{eq:reg_recourse}
\regrecourse(f;\smplst) = \sum_{i=1}^m \mathtt{sig}\left(-f(x)\right) \cdot \mathtt{sig}\left(-f(\dec_f(x))\right)
\end{equation*}
Where $\mathtt{sig} (z)=\frac{1}{1+e^{-z}}$ is the standard sigmoid function.
We calculate this regularization term using the same CCP approach on $\dec_f(x)$,
and once again in practice use $\decapx$ instead of $\dec$ in training.

\section{Experiments} \label{sec:experiments}

\begin{figure*}[t]
	\centering
	\includegraphics[width=0.27\textwidth]{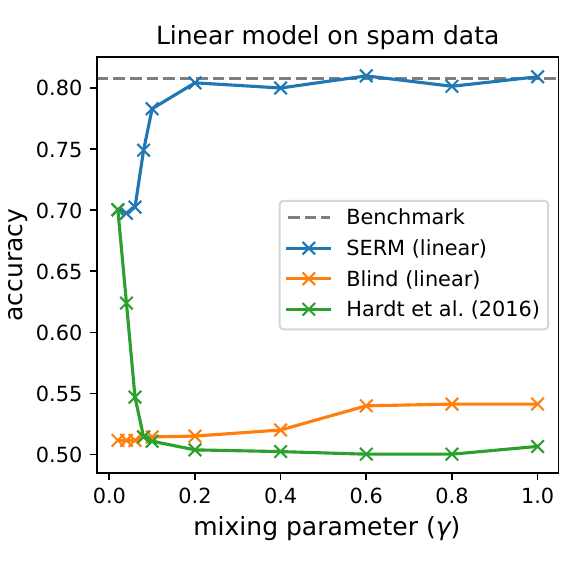} \,\,
		\includegraphics[width=0.27\textwidth]{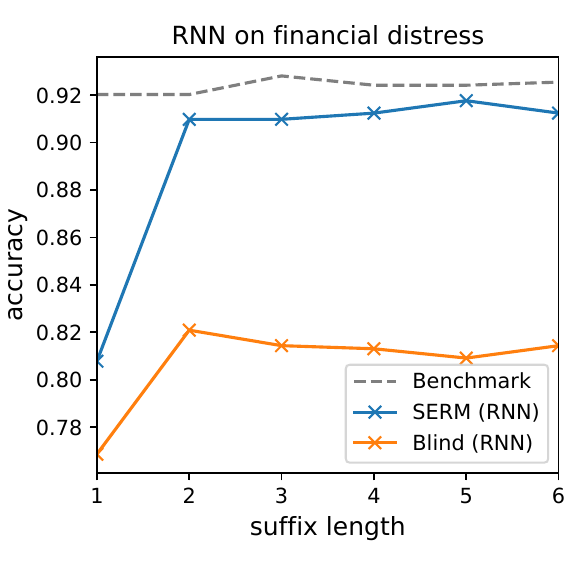} \,\,
	\includegraphics[width=0.36\textwidth]{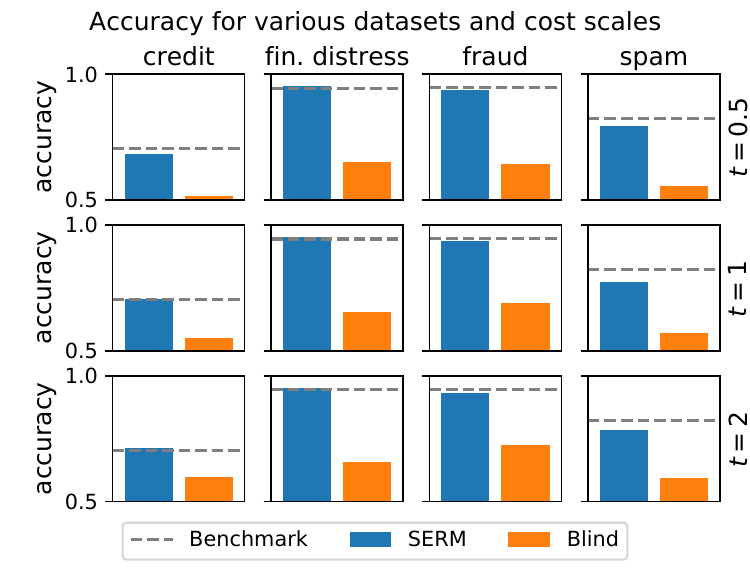} 
	\caption{A comparison of our \SERM\ approach to a `strategy-blind' baseline
	across multiple datasets, predictive models, and settings.
	\textbf{Left}: A reproduction of the setting of \citet{hardt2016strategic}
	on the spam dataset with mixed linear-quadratic cost.
	\textbf{Center}: Learning RNNs for financial distress time-series data.
	\textbf{Right}: Comparing across multiple datasets and degrees of gaming
	(controlled by cost scale $t$).}
	\label{fig:vanilla_rnn}
\end{figure*}

In this section we empirically demonstrate the utility and flexibility
of our approach on a diverse set of tasks and settings.
Our goal is to demonstrate how our framework supports learning in settings
that extend beyond the basic setting of strategic classification,
and each experiment extends the core setting of \citet{hardt2016strategic} 
in a way that targets a certain aspect of practical concern.
The Appendix includes further details, additional experiments, and illustrations.

\paragraph{Experimental setup.}
Across all experiments we use four real datasets:
(i) \spam\footnote{Data can be obtained from the authors of \citet{costa2014pollution}.},
	which includes features describing users of a large social network, some of which are spammers
	(used originally in \citet{hardt2016strategic});
(ii) \credit\footnote{\url{https://github.com/ustunb/actionable-recourse}},
which includes features describing credit card spending patterns, and labels indicating 
default on payment (we use the version from \citet{ustun2019actionable});
(iii) \fraud\footnote{\url{https://www.kaggle.com/mlg-ulb/creditcardfraud}},
which includes credit card transactions that are either genuine or fraudulent \citep{dal2015calibrating};
and (iv) \finance\footnote{\url{https://www.kaggle.com/shebrahimi/financial-distress}},
which includes time-series data describing businesses over time along with labels indicating their
level of financial distress and whether they have gone bankrupt.
All datasets include features that describe users 
and relate to tasks in which users have incentive to obtain positive predictive outcomes.
Some experiments use synthetic environments, described below.

We compare our approach (\SERM) to a strategy-blind baseline
that falsely assumes points do not move, achieved by training on the same
model class but using standard, non-strategic ERM.
When appropriate, we also compare to the strategic algorithm of \citet{hardt2016strategic},
and to other context-specific variants of our approach (e.g., \naive\ or oracle models).
All models train on non-strategic data, but are evaluated on strategic data.
As a benchmark, we use the strategy-blind model evaluated on non-strategic data
(i.e., where points do not move).

Our focus is mostly on linear classifiers as these are naturally interpretable
and hence justify the form of strategic behavior considered in strategic classification.
For time-series data, such as in experiment in Sec. \ref{sec:exp_rnn},
we use recursive neural networks (RNN).
For optimizing our approach we use Adam with randomized batches and early stopping
(most runs converged after at most 7 epochs).
User responses were simulated using $\dec$ from Eq. \eqref{eq:best_response}
with $\tau=1$ for training and $\tau=0.2$ for evaluation. 
Tolerance for CCP convergence was set to $0.01$ (most attempts converged after at most 5 iterations).
All methods use a 60-20-20 data split, and results are averaged over multiple random splits.


\begin{figure*}[t]
	\centering
	\includegraphics[width=0.33\textwidth]{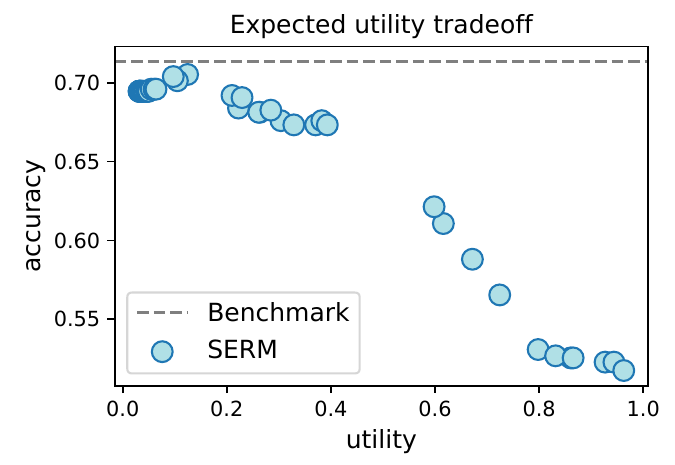}
	\includegraphics[width=0.33\textwidth]{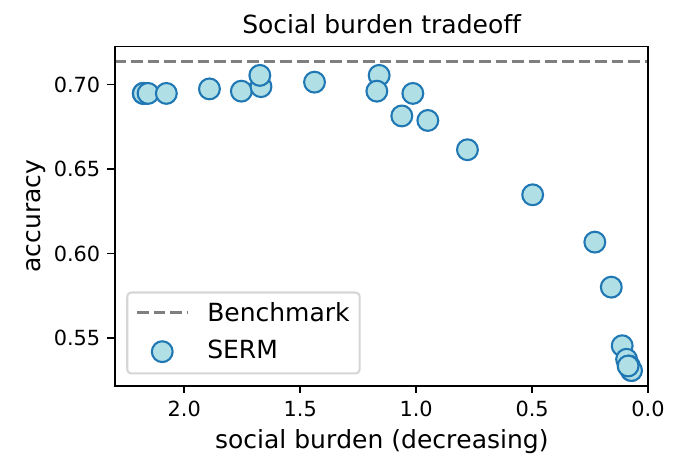}
	\includegraphics[width=0.33\textwidth]{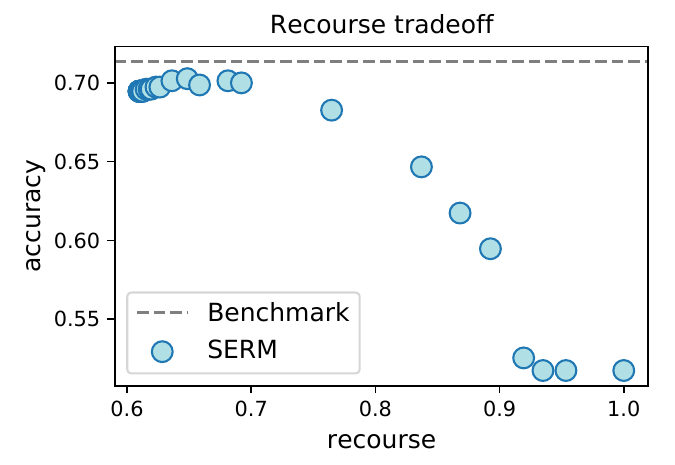}
	\caption{Trade-off curves of accuracy vs. various measures of social good:
		expected utility, social burden, and recourse.
		Points correspond to varying degrees of regularization ($\lambda$).
		For all measures, mild regularization improves social outcomes at little or no cost to accuracy.
	}
	\label{fig:reg}
\end{figure*}

%
%


\subsection{Core setting} \label{sec:exp_core}

Our first experiment begins with a reproduction of
the experimental setting of \citet{hardt2016strategic}.
They use the \spam\ dataset and consider linear-separable cost functions $\clin{\linvec}(x,x')=\max\{0,\linvec^\top (x'-x)\}$
with a specific hand-coded $\linvec \in \R^d$.
Their algorithm only supports separable costs,
and for linear-separable costs, returns a linear classifier.
Linear costs, however, are unstable under evaluation since points can move at zero cost
whenever $\linvec^\top (x'-x)\le 0$.\footnote{Linear costs are unstable at test time in
	the following way (see \citet{hardt2016strategic}, Fig. 2 caption):
	if $f$ is not precisely parallel to the cost vector $v$ (i.e., $f(x)=v^\top x+b$ for some $b$, where $c(x,x’)=\max\{0,v^\top (x’-x)\}$), then since any movement parallel to $v$ is free, any $x$ can move at zero cost to some $x’$ for which $f(x’)>0$ (and so all modified points are classified positively).}
Hence, they train with $\clin{\linvec}$,
but evaluate on a mixture cost:
\[
\cmix{}(x,x';\mix,\linvec)=(1-\mix) \cdot \clin{\linvec}(x,x')+\mix \cdot \cquad{}(x,x')
\]
where $\cquad{}=\|x'-x\|_2^2$ and $\mix \in (0,1]$ (they use $\mix \in \{0.1,0.3\}$).
Our approach supports general convex costs and so is able to learn using the `right' cost
function (i.e., that is used in evaluation) for any $\mix$.

Figure \ref{fig:vanilla_rnn} (left) shows performance for increasing $\mix$.
Benchmark accuracy on non-strategic data is 81\%,
but once strategic movement is permitted, performance of the strategy-blind model
drops to $\sim 54\%$ (data is balanced so chance $=\%50$).
The algorithm of \citet{hardt2016strategic} anticipates strategic behavior
but (wrongly) assumes costs are linear.
Accuracy for $\mix \approx 0$ improves to some extent ($\sim 69\%$), but remains far below the benchmark.
However, as $\mix$ increases, performance quickly drops to 50\%.
\tocut{\footnote{These results closely match those reported in \citet{hardt2016strategic}.}}
In contrast, by correctly anticipating strategic behavior,
\SERM\ restores accuracy almost in full for $\mix \ge 0.2$ ($\sim80\%$).


Next, we evaluate our method on additional datasets,
focusing on quadratic cost (as it does not require hand-coded parameters),
and varying the degree of `gaming' by scaling the cost by a factor of $t \in \{0.5,1,2\}$.
Figure \ref{fig:vanilla_rnn} (right) compares the performance of
SERM to the strategy-blind baseline. 
As can be seen, \SERM\ outperforms the blind model by a significant margin
and closely matches the non-strategic benchmark for all datasets and degrees of gaming.

\subsection{Social good regularization} \label{sec:exp_reg}
We evaluate our approach of regularizing for social good (Sec. \ref{sec:reg})
on \credit, which has been used in a recent paper on recourse by \citet{ustun2019actionable}.
Here we study how accuracy and social good trade-off under
\SERM\ by varying the amount of regularization $\lambda$.
Figure \ref{fig:reg} shows results for expected utility (left), social burden (center),
and recourse (right).
Each point in the plots corresponds to a model trained with a different $\lambda$.
Results show that all three measures of social good exhibit a super-linear tradeoff curve:
with mild regularization, a large increase in social good is obtained
at little or no cost to accuracy.
This highlights three interrelated points:
that incentives in strategic classification are not fully discordant;
that multiple models can achieve high accuracy;
and that of these, our approach to model selection through regularization
is effective at selecting models that promote favorable social outcomes.

\subsection{Beyond linear models} \label{sec:exp_rnn}
We apply our approach to time-series data in \finance\,
and for our predictive model we use non-linear recurrent neural networks (RNN) .
Each example in the dataset describes a firm over $t$ time steps using
a sequence of feature vectors $\x=(x^{(1)},\dots,x^{(t)})$,
where $x^{(k)} \in \R^d$ describes the firm at time $k\le t$ and $t$ varies across examples ($1 \le t \le 14$).
Labels determine whether the firm has gone bankrupt at time $t$,
and we assume users can modify features only at this final timepoint $x^{(t)}$.
This mimics a setting in which the history of the firm is fixed,
but its features at the time of arbitration can be manipulated (at some cost).
We implement $f$ as a fully-connected RNN with layers:
\[
h^{(i)} = \varphi(W x^{(i)} + V h^{(i-1)} + a)
\]
where model parameters $W,V,a$ are tied across layers
and activations $\varphi$ are sigmoidal for all layers but the last, which is used for prediction.
We set the embedded dimension to $10$.
Plugging into Eq. \eqref{eq:best_response}, the response mapping becomes:
\[
\dec(\x) = \argmax_{x' \in \X} \, \sign(w^\top x' + v^\top h^{(t-1)} + b) - c(x^{(t)},x')
\]
with quadratic cost.
This is a special case of Sec. \ref{sec:nonlinear} with $\rep(x)=(x^{(t)},h^{(t-1)})$.
Since $f$ is linear in $x'$ and $c$ is convex in $x'$,
our CCP approach can be applied (Sec. \ref{sec:diff_through_response}).

We experiment with varying history lengths by taking suffixes of fixed size $k$ of each $\x$.
Figure \ref{fig:vanilla_rnn} (center) shows results for increasing suffix lengths $k$.
As can be seen, \SERM\ outperforms a blind RNN by roughly $15\%$ for all $k$
and nearly matches the non-strategic benchmark for $k \ge 2$.

\begin{figure*}[th!]
	\centering
	\includegraphics[width=\textwidth]{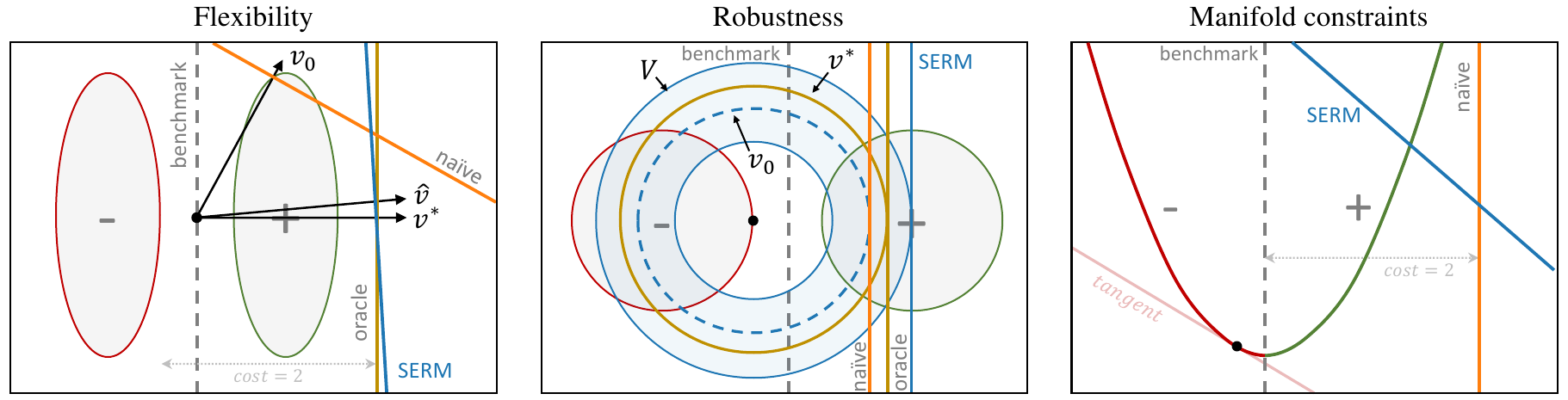}
	\caption{An illustrations of various extensions of the role of the cost function $c$.
	\textbf{(Left)} \textit{Flexibility}: cost is linear (arrows show direction)
	and is initially $\linvec_0$, but the system 
	can invest effort to change it. Learning produces a cost $\hat{\linvec}$ that is near the optimal $\linvec^*$, with which \SERM\ achieves near-optimal accuracy.
	\textbf{(Center)} \textit{Robustness}: Cost is quadratic (circles show diameter of maximal movement),
	but the system has only an estimate $v_0$ of the true cost $v^*$.
	Learning aims to be minmax-optimal with respect to all costs $v \in V$ near $v_0$
	(ring shows the set $V$). \SERM\ again learns a near-optimal classifier.
	\textbf{(Right)} \textit{Manifold constraints}: Points are allowed to move only
	on the quadratic manifold that is unknown (and hence not encoded in $c$).
	Our approach is to first learn the manifold, and then use derived tangents as constraints
	in optimizing $\dec$.
	This approximates movement within the manifold and allows \SERM\ to learn well.
}
	\label{fig:costs}
\end{figure*}

\subsection{Beyond oracle cost functions} \label{sec:exp_costs}
In this section we move beyond the assumption of a fixed and known (`oracle') cost function
and study the extensions presented in Sec. \ref{sec:extensions}.
We use stylized 2D synthetic data that allows us
to visually illustrate how our approach behaves in these extended settings.
In each setting we compare \SERM\ to appropriate \emph{\naive} and \emph{oracle}
baselines, specified below.

\paragraph{Flexible cost functions.}
In this setting we assume that the system can modify an initial cost to some extent.
Data is generated by a mixture of two `narrow' MVN distributions
with means $\mu=[-0.5,0]$ (for $y=-1$) and $\mu=[0.5,0]$ ($y=1$)
and diagonal covariances with $\Sigma_{11}=0.1$ and $\Sigma_{22}=1$.
We use linear-separable cost $\clin{\linvec}$ (for stability we mix using $\mix=0.005$)
and set the initial cost to $\linvec_0=[0.5,0.5]$.
The learner has flexibility to choose any
$\linvec \in V=\{\linvec' : \|\linvec'-\linvec_0\|_\infty \le 2 \}$.
For a \naive\ baseline we consider a model that does not modify the cost, i.e., uses $\linvec=\linvec_0$.
The oracle baseline uses the optimal $\linvec^* = [2,0] \in V$.

To learn with flexible costs,
we use the approach presented in Sec. \ref{sec:flexibility}
and jointly optimize model parameters $w,b$ and cost parameters $\linvec$.
Results are shown in Figure \ref{fig:costs} (left).
If no strategic movement is allowed, then the optimal linear classifier having
$w=[1,0],b=0$ achieves $\sim 100\%$ (non-strategic) accuracy.
However, with strategic movement under $\linvec_0$,
the \emph{strategically-optimal} classifier has weights $w=[0.5,0.5],b=1$.
These are indeed the weights learned by the \naive\ baseline, which achieves
0.72\% accuracy.
However, for $\linvec=[1,0]$, the optimal classifier has $w=[1,0], b=1$,
and an oracle baseline that has $\linvec^*=[1,0]$ learns these weights and achieves 100\%.
Our flexible approach learns a near-optimal cost ($\hat{\linvec} = [2.3,0.6]$)
along with the corresponding optimal model, 
achieving 97\% accuracy.



\paragraph{Unknown cost functions.}
In this setting we assume that points move according to a ground-truth cost function
that is unknown, and that the system has only an rough estimate of this cost.
Data is generated by a mixture of two symmetric MVN distributions
with means $\mu=[-0.6,0]$ (for $y=-1$) and $\mu=[0.6,0]$ ($y=1$)
and diagonal covariances $\Sigma_{11}=\Sigma_{22}=0.1$.
We use an a-symmetric quadratic cost in which dimensions can differ in scale,
$\cquad{v}(x,x') = \sum_{i=1}^d v_i (x_i-x'_i)^2$,
parameterized by $v \in \R^d$.
The ground-truth cost is set to $v^*=(0.5,0.5)$ and the initial estimate to $v_0=(2,2)$.
Here the oracle baseline learns using $v^*$,
and \naive\ baseline learns under the false assumption that $v_0=v^*$.

To learn under an unknown cost,
we use the robust minmax approach presented in Sec. \ref{sec:robustness}
and train a predictor $f$ to be minmax-optimal w.r.t. a belief set $V$,
which we set to
$V = \{v \in \R^2 : \|v_0-v \|_\infty \le 1.7 \}$
(note that $v^* \in V$).
In this setting 
optimization is simplified since it suffices to maximize $v$ over the reduced set
$\{v_{\text{min}}, v_{\text{max}}\} \subset V$, where in our case
$v_{\text{min}}=[0.3,0.3]$ and $v_{\text{max}}=[3.7,3.7]$.
Figure \ref{fig:costs} (center) illustrates the results of learning in this setting.
The benchmark on non-strategic data is $\sim 100\%$, as is the performance
of the oracle model on strategic data.
The \naive\ model obtains 74\% accuracy using $v_0$.
SERM\ optimized using the minmax objective achieves $91\%$ accuracy.

\paragraph{Manifold constraints.}
In this setting we work with a cost function is known and fixed,
but assume that points can move only within a low-dimensional manifold that is unknown to the system.
We use a 1D manifold that satisfies $x_2=-x_1^2$ and generate data points $x=(x_1,x_2)$
using $x_1 \sim U([-5,5])$ and set $x_2$ according to the manifold constraints.
Labels are $y=\sign(x_1)$ and the cost is quadratic.

To learn under unknown manifold constraints, we follow the approach outlined in Sec. \ref{sec:manifold},
where in the response function $\dec$ the (unknown) manifold constraints
are approximated using (estimated) tangent constraints.
Here, manifold tangents have a close form solution, but for completeness
we pursue a more general approach and learn them.
In particular, we learn the manifold using a contractive autoencoder (CAE) \citep{rifai2011contractive}
with 2 stacked hidden layers of width 20 and sigmoidal activations.
CAEs minimize reconstruction error under a regularization term
that penalizes Jacobian absolute values.
Balancing between these two forces encourages learned embeddings that permit movement
only in directions that are useful for reconstruction, i.e., along the manifold.
Jacobians also provide manifold tangents, and we compute these as proposed
for manifold tangent classifiers (MTC) in \citet{rifai2011manifold}.

Figure \ref{fig:costs} (right) shows results for this setting.
The problem is separable and so optimal non-strategic accuracy is 100\%.
A \naive\ baseline which does not account for manifold constraints
wrongly assumes points can move freely across the $x_1$-axis and places all weight on this axis.
This false assumption reduces accuracy to 52\%---much lower than
that of a blind baseline, which achieves 89\%.
Our \SERM\ approach augmented with MTC tangents over an estimated CAE manifold reaches 98\% accuracy.

\section{Conclusions}
In this paper we proposed a practical approach to learning in strategic classification.
By differentiating through user responses, our approach allows for
effective and efficient learning in diverse settings.
Key to our approach were differentiable optimization solvers;
we are hopeful that future advances in this field could be utilized 
within our framework to support more elaborate forms of user responses.
Strategic classification has so far been mostly studied under a theoretical lens;
our work takes a first step towards making learning practical,
with the aim of promoting discussion amongst practitioners,
applied researchers, and those interested in social aspects of learning.
Our approach to regularization serves as a reminder that strategic classification is a game for two players,
and that learning can, and should, aid in promoting the interests of all parties involved.
We believe that a responsible approach to applied strategic classification
can be beneficial to both systems and the users they serve.


%
%
%

	\bibliography{refs}
	\bibliographystyle{icml2021}
	

\appendix
\onecolumn

\section{Experimental details}
\subsection{Data}

Across all experiments we use four real datasets:
\begin{itemize}
\item 
\textbf{\texttt{Spam}}\footnote{Data can be obtained from the authors of \citet{costa2014pollution}.}.
This dataset includes features of authentic users and spammers from a large social network, and was used in \citet{hardt2016strategic}.
The data includes $n=7,076$ examples and $d=15$ features.
Features include number of words in the post, number of phone numbers in the post and number of followers of the user.
The data is balanced, i.e., there is an equal number
of positive ($y=1$) and negative ($y=-1$) examples.

\item
\textbf{\texttt{Credit}}\footnote{\url{https://github.com/ustunb/actionable-recourse}}.
This dataset includes features describing credit card spending patterns,
along with labels indicating default on payment.
We use the same version used in the recourse paper by \citet{ustun2019actionable},
and adopt their preprocessing procedure (see link).
The original dataset includes $n=30,000$ examples and $d=11$ Features.
Features include age, amount of bill statement and history of past payments.
For training time purposes,
in our experiments we used a random balanced subset of 3,000 examples,
as we did not notice any changes in performance for larger sample set sizes
(which is plausible given that $f$ is linear and $d$ here is small).

\item
\textbf{\texttt{Fraud}}\footnote{\url{https://www.kaggle.com/mlg-ulb/creditcardfraud}} \citep{dal2015calibrating}.
This dataset includes credit card transactions that are either genuine or fraudulent.
The original data includes $n=284k$ examples and $d=29$ features.
Due to confidentiality issues, feature information is not provided.
The original dataset is highly imbalanced, and for our experiments we take all
available negative examples and uniformly sample a matching number of positive examples, giving a total of $n=984$ balanced examples.

\item
\textbf{\texttt{Finance}}\footnote{\url{https://www.kaggle.com/shebrahimi/financial-distress}}.
This dataset includes time-series data describing firms along with an indication of their level of financial distress (denoted $\mathtt{fd}$).
Each series ends either at the maximal time step of $t=14$
or earlier if the firm has gone bankrupt.
Bankrupcy is declared when $\mathtt{fd}<-0.5$, and we use this
definition to determine labeles $y$.
The data includes $n=422$ time-series examples, with each time step
within each example described using $d=83$ anonymized numerical features.\footnote{The data includes a single categorical feature, which we discard.}
The ratio of positive (i.e., non-bankrupt) examples is 67.7\%.
Time-series lengths are in the range $t \in \{0,\dots,14\}$ (inclusive),
with mean=8.7 and median=4.9.
\end{itemize}
Features in all datasets were standardized
(i.e., scaled to obtain a mean of zero and standard deviation of one)
on the train set.
To ensure consistent behavior in term of the effect of the cost
function on movement, features in each datasets
were further divided by $\sqrt{d}$ where $d$ is the (per-dataset)
number of features.




\subsection{Training and tuning} \label{app:train_tune}

For all experiments we use a 60-20-20 split into train, validation, and test sets, respectively,
and all results are averaged over multiple random splits.
For optimization we use ADAM,
where within each epoch use randomize batches of size
24 for \finance\ and \fraud\,
64 for \credit\, and 128 for \spam\,
chosen according to their respective number of examples,
and early stop w.r.t. accuracy on the validation set.
For all methods, learning rates were set according to the validation set.
For $\sigma$ we set $\tau=1$ for training and $\tau=0.2$ for testing,
and set the CCP tolerance to 0.001,
but note results are robust to variations in these.

\subsection{CCP procedure}
We use our own implementation of CCP, based on
\citet{shen2016disciplined},
and using the publicly-available SCS solver\footnote{\url{https://web.stanford.edu/~boyd/papers/scs.html}}
for solving the concave sub-problems appearing in each iteration.
Algorithm \ref{algo:fwd} includes pseudocode for our procedure.
We use the notation:
\[
\sigma_\cup(x;f) =  \frac{1}{2} \sqrt{\left(\temp^{-1} f(x) +1\right)^{2}+1},
\qquad
\sigma_\cap(x;f) = - \frac{1}{2} \sqrt{\left(\temp^{-1} f(x) - 1 \right)^{2}+1},
\]
for the convex and concave parts of $\sigma$, respectively.
 \begin{algorithm}[H]
 	\caption{{CCP forward pass}}
 	\begin{algorithmic}[1]
 		\STATE{Initialize $t=0, \delta=\infty$}
 		\STATE{$x^0 = x$}
 		\REPEAT
 		\STATE{$t=t+1$}
 		\STATE{$g = \nabla_x \sigma_\cup(x^{t-1};f)$} 
  		\hfill \COMMENT{linearized convex term at $x^{t-1}$}
 		\STATE{$x^t = \argmax_{x'} g^\top x' + \sigma_\cap(x';f) - c(x,x')$}
	 		\hfill \COMMENT{get argmax of current concave proxy}
 		\UNTIL{convergence}
		\STATE{$g^t = \nabla_x \sigma_\cup(x^{t};f)$} 
	 		\hfill \COMMENT{final linearized term, to be used for backward pass}
 		\STATE{\textbf{return} $x^t,g^t$}
 	\end{algorithmic}
 	\label{algo:fwd}
 \end{algorithm}
In practice we terminate when $\|x^t-x^{t-1}\|_2 \le \mathtt{tol}$
or after at most $t=100$ iterations.
In our experiments we used $\mathtt{tol}=0.001$, but the vast majority
of instances terminated after $t=5$ iterations.
The algorithm returns two objects:
the approximate argmax $x^t$ used in the forward pass,
and the linearization $g^t$ (of which $x^t$ is an exact argmax)
used to parameterize the convex optimization layer to enable a backwards pass.



\subsection{Runtime evaluation}
In this section we present results on the runtime of our method.
Presumably the main overhead in our approach is computing
the CCP solution and surrogate in the forward pass.
Note however that while each call to solver may be expensive,
jointly solving for all examples in a batch
(i.e., solving $k$ independent problems in one call,
where $k$ is the batch size) can greatly reduce this overhead.
All experiments were run on a single laptop
(Intel(R) Core(TM) i7-7500U CPU @ 2.70GHz 2.90 GHz,
16.0 GB RAM).

\begin{figure}[b!]
	\centering
	\includegraphics[width=0.24\textwidth]{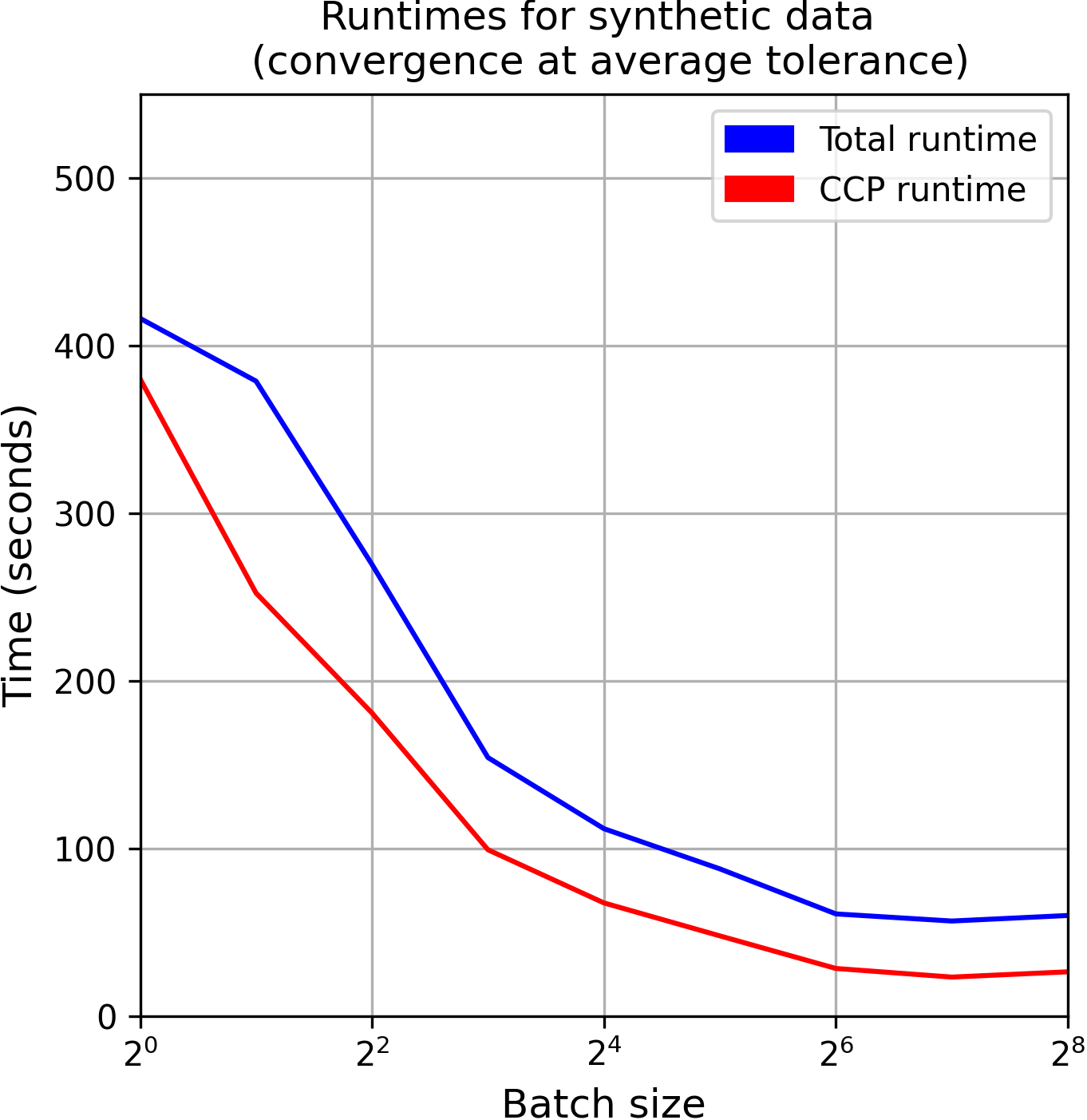}
	\includegraphics[width=0.24\textwidth]{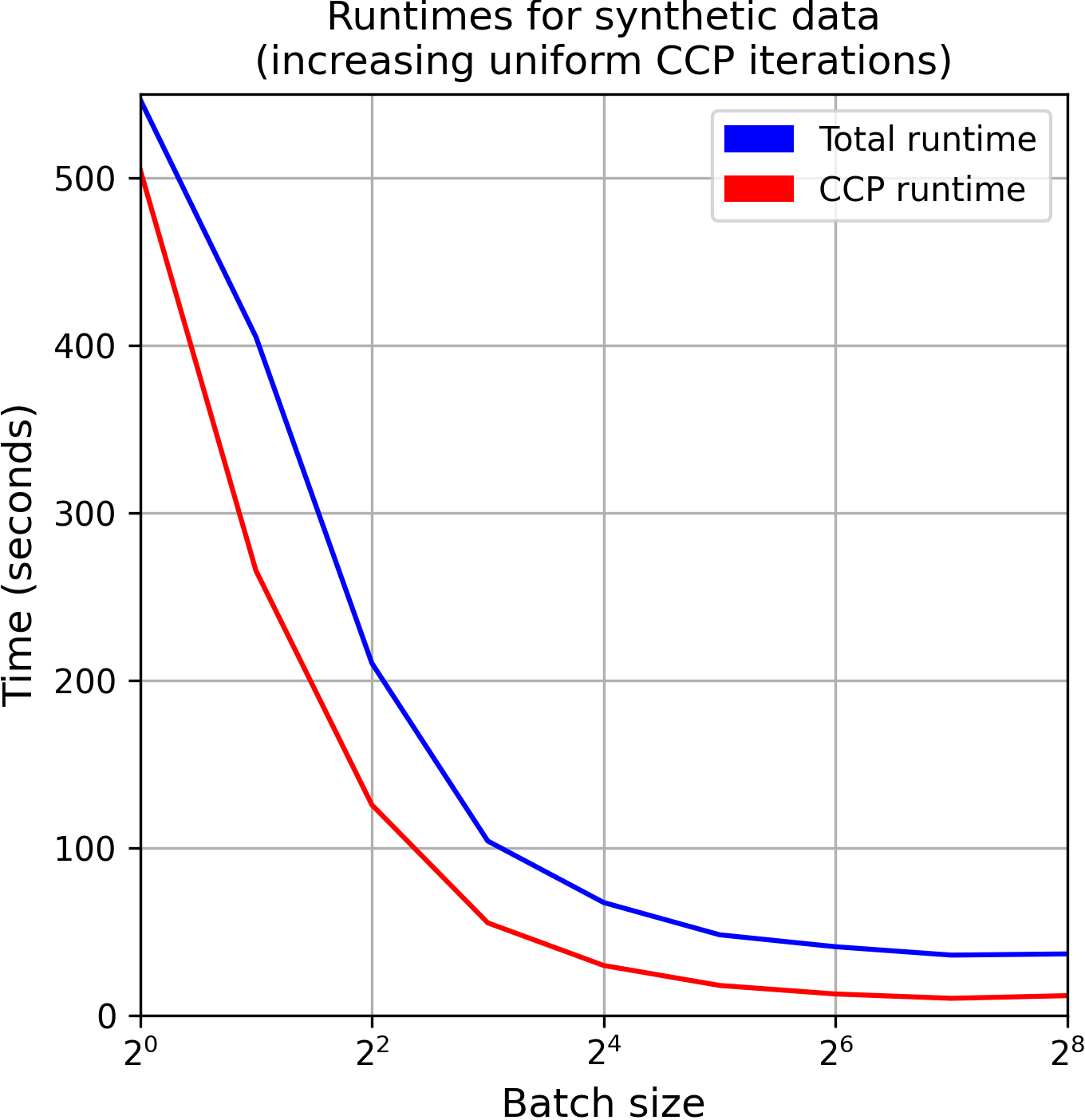}
	\includegraphics[width=0.24\textwidth]{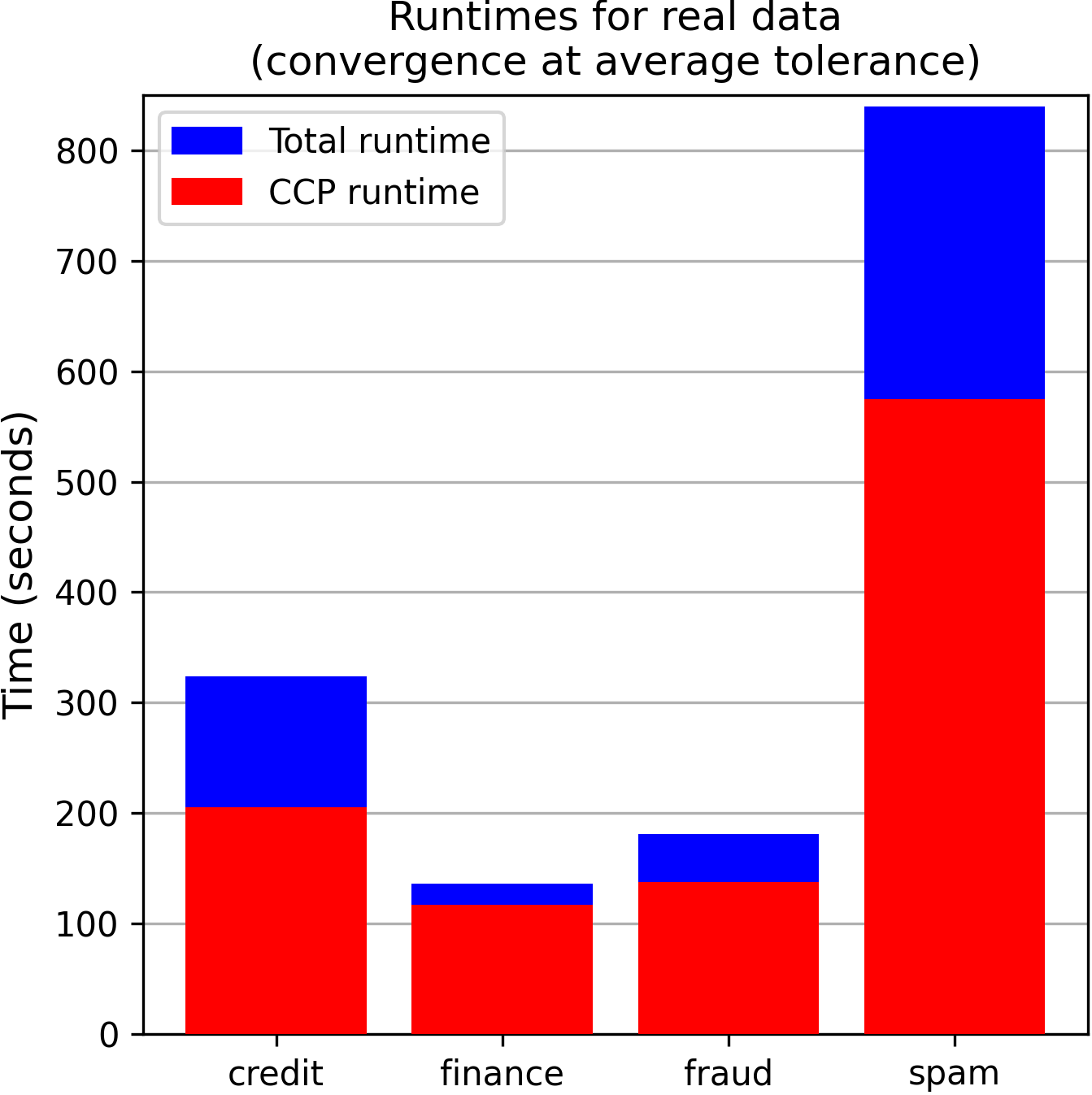}
	\includegraphics[width=0.24\textwidth]{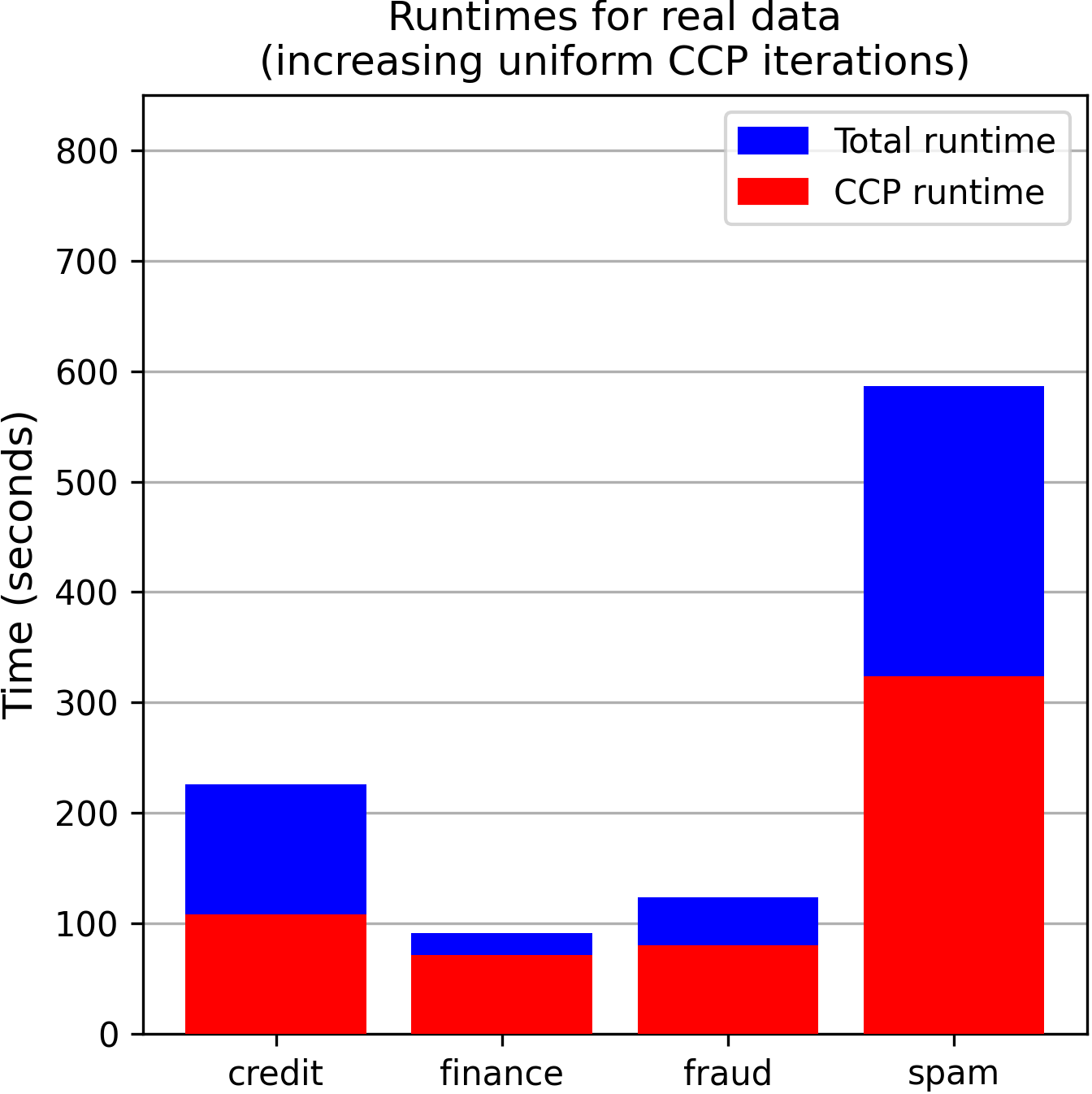}
	\caption{Runtime on synthetic data (varying CCP batch size)
	and real data (fixed batch size).
	}
	\label{fig:runtime}
\end{figure}

To evaluate this, we run an experiment on synthetic data
and for varying batch sizes.
We use scikit-learn's \texttt{make\_classification} function, 
using 750 train samples, 250 validation samples,
$d=5$ features, balanced classes, and 0.01 label noise.
\ref{fig:runtime} (two left-most plots) show runtime results
for increasing batch size $k$ and training for 5 epochs.
We consider two alternatives to setting the number of CCP iterations:
(i) convergence to tolerance 0.001 on average across examples in the batch (left), and
(ii) a slowly increasing, uniform number of iterations,
with the number of iterations initialized at one
and increased by one after each epoch (right).
As can be seen, increasing the batch size
greatly improves overall runtime,
as well as the the relative runtime of CCP.

We also report runtime on the real datasets used in Sec. \ref{sec:exp_core}.
For equalized comparison we set the number of epochs to 7.
Batch sizes were set per dataset as described in Sec. \ref{app:train_tune} (as noted,
these were not chosen to optimize for speed).
Figure \ref{fig:runtime} (two right-most plots) show total train times and relative CCP times for a all datasets
for both stopping criteria.

\paragraph{Code.}
Our code can be obtained at:
\url{https://github.com/SagiLevanon1/scmp}.


\end{document}